\documentclass[11pt]{article}

\usepackage[margin=1in]{geometry}
\usepackage{graphicx}
\usepackage{booktabs}
\usepackage{float}
\usepackage{amsmath,amssymb}
\usepackage{natbib}
\usepackage{authblk}
\usepackage{caption}
\usepackage{subcaption}
\usepackage{xcolor}
\usepackage{siunitx}
\usepackage[colorlinks=true,linkcolor=blue!55!black,citecolor=blue!45!black,urlcolor=blue!55!black]{hyperref}
\usepackage{microtype}

\setcitestyle{authoryear,round}
\newcommand{\aion}{AION-1}
\newcommand{\dmag}{|\Delta\mathrm{mag}|}
\newcommand{\dzreq}{|\Delta\langle z\rangle|/(1+z)}

\title{\bfseries A survey detection channel overrides the pixels in an astronomical
foundation model,\\ and biases tomographic mean redshifts}

\author[1]{Ihor Kendiukhov}
\affil[1]{Independent researcher}

\date{\today}

\begin{document}
\maketitle

\begin{abstract}
\noindent
Foundation models for astronomy are trained on survey pixels together with the catalogue
products derived from those pixels. Those catalogues are incomplete at a measurable rate, and we
show that a model trained on both inherits that incompleteness as a physical systematic.
\aion{} \citep{parker2025aion}---a 39-modality transformer trained on more than
$2\times10^{8}$ objects from five surveys and released with open weights---is the instance we
audit, using causal interventions on its inputs.

Our central finding is that \aion{} \emph{defers to catalogue-derived metadata in preference to
the raw pixels, and does not discount that metadata when it contradicts them}. Holding image
pixels byte-identical and altering only the survey segmentation map changes every quantity the
model reports---flux, size, ellipticity, redshift---by 110--4400 times a matched placebo.
Displacing the mask by \SI{25}{px} reduces the $g$-band flux estimate by 79\,\% although every
photon remains in the image. The mechanism is detection \emph{gating} rather than masked
aperture photometry: the response tracks whether the mask asserts a source at the field centre
($r=0.47$) more strongly than the light the mask encloses ($r=0.30$), and across 322 severely
blended systems the model ignores entirely how the pipeline partitioned the light
($R=-0.006\pm0.001$). The behaviour is not confined to the detection channel: corrupting the
catalogue photometry moves the $g$-band readout by \SI{0.61}{mag}, leaving the model
\emph{nine times worse than supplying no metadata at all}.

We quantify the consequence. The Legacy Survey pipeline \citep{dey2019legacy} leaves 3.68\,\% of
targets with no segment covering their position, measured over 5000 cutouts. Representing such a
miss by the segmentation fields the pipeline actually returns when it misses, and propagating
that measured rate through \aion{}'s photometric redshifts, shifts the mean redshift of
tomographic bins by a median 0.71 times the LSST DESC requirement \citep{lsstdesc2018srd} over
40 miss assignments and exceeds it in 12 of them---one routine failure mode consuming most of
the error budget, and occasionally all of it. Drawing the missed objects by their measured
magnitude-dependent miss probability rather than uniformly leaves this unchanged within the
realisation scatter. Adding
positional errors of the observed magnitude takes the worst bin to 8.3 times the requirement,
consistent with our finding that the model is far more sensitive to \emph{where} the mask is
than to its shape.
Supplying spectra removes the effect almost entirely, so the systematic is specific to
photometric-only inference---the regime in which such a model is most attractive. Withholding
the detection channel eliminates it at no measurable cost, whereas substituting a generic mask
is substantially worse than either.

Two further limitations live in the tokeniser rather than the transformer. Measured over
$1.15\times10^{6}$ tokens, the image codec emits 8.3\,\% of its 4375 codes, and on patches
containing the source its effective vocabulary is 28 codes against 934 of 1024 for the spectrum
codec; those 28 states form a one-dimensional brightness ladder. The redshift readout is
quantisation-limited rather than information-limited. Codebook utilisation is unchanged at
$2.7\times$ the parameter count, so both persist under scaling. Finally, sparse dictionaries are unreliable as causal
handles on the gate: across 15 dictionaries spanning width, sparsity and seed, recovery spans
26--75\,\% and moves by up to 18 points on the seed alone, thirteen of fifteen lose to
difference-in-means, and reconstruction quality does not predict which will steer.
\end{abstract}

\section{Introduction}
\label{sec:intro}

Foundation models have arrived in astronomy. Following cross-modal and self-supervised models
for galaxies \citep{parker2024astroclip,smith2024astropt,rizhko2025astrom3,leung2024stars} and
the assembly of large heterogeneous corpora \citep{mmu2024multimodal}, and cross-domain scientific pretraining more broadly \citep{mccabe2023mpp,mccabe2025walrus}, \aion{}
\citep{parker2025aion} trains a single encoder--decoder transformer over discrete tokens
spanning 39 observational modalities from the DESI Legacy Imaging Surveys
\citep{dey2019legacy}, Hyper Suprime-Cam \citep{aihara2018hsc,aihara2022hscpdr3}, SDSS
\citep{york2000sdss}, DESI \citep{desi2016experiment,abareshi2022desiinstrument,desi2024edr,desi2025dr1} and Gaia
\citep{gaia2016mission,gaia2023dr3}. It is released openly at three scales, and it reaches or
exceeds task-specific baselines across many downstream problems using a frozen encoder and a
light probe head.

The case against deploying such a model inside a cosmological analysis is not accuracy but
accountability. No survey collaboration places a learned representation in a likelihood without
a correlated error budget, and ``what is the $R^2$'' is never the blocking question. The
blocking question is what the model actually uses, and whether that introduces a spatially or
systematically coherent bias. Photometric redshift requirements, in particular, constrain the
\emph{mean} redshift of a tomographic bin rather than per-object scatter
\citep{lsstdesc2018srd,newman2022photoz}, so a small coherent shift matters far more than a
large random one.

Mechanistic interpretability offers tools for this: linear probes
\citep{alain2016understanding,belinkov2022probing}, activation patching and causal tracing
\citep{vig2020causal,meng2022locating}, concept erasure \citep{ravfogel2020null,belrose2023leace},
and sparse dictionary learning
\citep{cunningham2023sparse,gao2024scaling,bussmann2024batchtopk,rajamanoharan2024jumping},
together with its transcoder and crosscoder variants \citep{dunefsky2024transcoders,lindsey2024crosscoders}.
These were developed largely for language models, where the superposition hypothesis
\citep{elhage2022toy} and the linear representation hypothesis \citep{park2024linear} frame the
questions. Their transfer to scientific models is an active and, so far, mixed story: sparse
autoencoders have been applied to galaxy morphology
\citep{wu2025euclid,wu2025sparsefeature}, protein language
models \citep{simon2025interplm}, single-cell models \citep{pedrocchi2025singlecell} and vision
encoders \citep{zaigrajew2025clip}, while an interpretability study of a sibling
continuum-dynamics foundation model found features that were only piecewise consistent and
matched no standard physical basis \citep{rosenfeld2026sparse}.

\aion{} is an unusually tractable target, for reasons specific to its design:
\begin{enumerate}\itemsep2pt
\item \textbf{Token identifiers decode to physical quantities.} Most modalities are single-token
scalars quantised to 1024 centroids by a parameter-free empirical-CDF codec, so attributions can
be reported in magnitudes or dex rather than in logits.
\item \textbf{Token index is a physical coordinate.} Images are a $24\times24$ grid of
$4\times4$-pixel patches at known sky positions; spectra are 273 tokens on a known wavelength grid.
\item \textbf{The decoder query carries no value information.} For a single-token target the
query is effectively a constant vector, so the readout measures what the encoder supplied.
\item \textbf{A controlled experiment sits in the checkpoint.} HSC and Legacy imaging share one
codec and one 4375-code vocabulary but have independent embedding tables.
\item \textbf{Uniquely among public astronomical foundation models, \aion{} ingests a detection
map}---the survey's segmentation image---as an input modality. The pipeline's own decisions are
therefore directly editable, which is the basis of most of what follows.
\end{enumerate}

The behaviour we find, however, is not specific to astronomy. It follows from a design choice
that multimodal scientific foundation models make routinely: training on raw observations
\emph{together with} catalogue products that were themselves derived from those observations.
Such a channel is close to a free answer during pretraining, and nothing in a masked-modelling
objective penalises a model for preferring it. \aion{} lets us measure the consequence exactly,
because one of its metadata channels---the detection map---is an image we can edit while holding
the pixels byte-identical.

Our contributions are:
\begin{enumerate}\itemsep2pt
\item A causal demonstration that \aion{} prefers a catalogue-derived channel to the raw pixels,
and fails to discount it under contradiction, across every quantity it reports
(\S\ref{sec:gate}). We isolate the mechanism as detection \emph{gating} rather than masked
aperture photometry, and show on real blended systems that deblending decisions do not propagate
while positional ones do.
\item Evidence that the preference is a property of the design pattern rather than of one
channel: contradicted catalogue \emph{photometry} leaves the model nine times worse than
supplying no metadata at all (\S\ref{sec:general}), and the vulnerability grows with model
scale (\S\ref{sec:impact} onwards).
\item A quantified downstream cost, propagated from the survey's own measured failure rate
rather than an assumed one, together with a full statement of the assumptions behind that
estimate (\S\ref{sec:impact}), and a mitigation that removes the vulnerability at no measurable
cost.
\item Two limits attributable to the tokeniser rather than the transformer, neither of which
scaling addresses (\S\ref{sec:tok}).
\item A measurement of how unreliable sparse dictionaries are as causal handles, over 15
dictionaries: the seed-to-seed spread exceeds the gap to the linear baselines, and
reconstruction quality does not predict causal utility, so a single-seed evaluation cannot
settle the question either way (\S\ref{sec:sae}).
\end{enumerate}

We use causal interventions throughout rather than attention inspection, because twelve layers
of bidirectional self-attention make attention maps unusable as attribution here. Every effect
is reported against a matched placebo, and every point estimate carries a bootstrap interval.

\section{Model, data and methods}
\label{sec:methods}

\paragraph{Model.} \texttt{aion-base} (\num{314.3}\,M parameters; 12 encoder and 12 decoder
blocks, $d=768$) for all experiments, with \texttt{aion-large} (\num{859.7}\,M; 24/24,
$d=1024$) for the scaling test. \aion{} follows the 4M multimodal masked-modelling recipe
\citep{mizrahi20234m,bachmann20244m21}, itself built on masked image modelling
\citep{he2021masked,chang2022maskgit} and the transformer and ViT architectures
\citep{vaswani2017attention,dosovitskiy2020image}. Images are tokenised with finite scalar
quantisation \citep{mentzer2023finite} over a MagViT-style backbone
\citep{yu2024language,esser2020taming}, spectra with a lookup-free quantiser over a ConvNeXt-V2
encoder \citep{liu2022convnet,woo2023convnextv2}, described separately by
\citet{shen2025spectral}, in the VQ-VAE tradition
\citep{vandenoord2017neural,yu2022vector,huh2023straightening}.

\paragraph{Data.} (i) A \SI{113}{GiB} cross-match of \num{115404} galaxies with DESI spectra,
Legacy Survey $grz$+WISE imaging with per-object segmentation maps, and PROVABGS physical
parameters \citep{hahn2023provabgs,hahn2023bgs}; photometry and segmentation derive from the
\texttt{legacypipe}/Tractor pipeline \citep{lang2016tractor,lang2025legacypipe}, whose detection stage
thresholds SED-matched-filter detection maps at $6\sigma$ and groups the surviving peaks into
blobs that are jointly forward-modelled, rather than deblending by isophotal segmentation
\citep{bertin1996sextractor}. (ii) One
\SI{1.9}{GB} shard of MultimodalUniverse HSC PDR3 deep/ultradeep \citep{mmu2024multimodal,
aihara2022hscpdr3}, 1883 galaxies, for the HSC codebook measurement. (iii) The \aion{} authors'
own test fixtures, used as regression tests.

\paragraph{Sampling.} All samples are drawn at random from the full catalogue. This matters: the
file is \textsc{healpix}-ordered, and an initial analysis using the first $N$ rows---a single
contiguous $2.3^\circ\times2.5^\circ$ field---gave a redshift probe $R^2$ of 0.48 where a
random draw gives 0.91.

\paragraph{Interventions, in plain terms.} Two operations recur below, and neither assumes
familiarity with the interpretability literature.

The first is an \emph{input} intervention: we alter one input channel---usually the segmentation
map---while holding every other input byte-identical, and record how the quantity the model
reports moves. Each such effect is quoted against a \emph{matched placebo}: an edit of
comparable size to a channel that should not matter, here displacing the right-ascension token
by 250 codebook steps. The placebo is what rules out the possibility that merely perturbing an
input moves the answer.

The second is an \emph{internal} intervention. The encoder holds each object as a vector of
$d=768$ numbers at each of its twelve blocks. We add a fixed vector $v$ to that representation
at one block and let the remaining blocks run---the operation the interpretability literature
calls \emph{steering}. The $v$ we add is simply the difference between the average internal
state under a true mask and under a displaced one, a difference in means. If that direction is
what carries the gate, adding it back should undo the damage a wrong mask does. We report
\emph{recovery}: the fraction of the gate-induced collapse restored, where 0 is the displaced-mask
value and 1 the true-mask value. Every candidate direction is rescaled to the same length before
it is added (\emph{matched norm}), so the coefficient $\alpha$ means the same size of
intervention whichever method produced the direction, and a random vector of that same length is
always run as a control.

\paragraph{Readouts.} Unless stated otherwise, scalar quantities are read as the \emph{mean} of
the token posterior. The exceptions, where the \emph{median} is used, are
Table~\ref{tab:tomo}, the redshift-precision figures of \S\ref{sec:tok} and all of
\S\ref{sec:calib}; \S\ref{sec:calib} shows why the median is the better estimator for
\texttt{tok\_z}, and we recommend it. We did not recompute the intervention tables with it
because they compare arms measured on the same objects with the same estimator, so the
\emph{differences} they report are unaffected; the one place this matters is the absolute
redshift error in Table~\ref{tab:mitig}, which is therefore larger than a median readout would
give and should be read only across rows. Flux errors are reported in
magnitudes, $\dmag = 2.5\,|\log_{10}(\hat f / f)|$, which is bounded and unaffected by the small
denominators that make fractional error unusable for faint bands.

\section{The detection gate}
\label{sec:gate}

\subsection{Existence}

\begin{figure}[t]
\centering
\includegraphics[width=\textwidth]{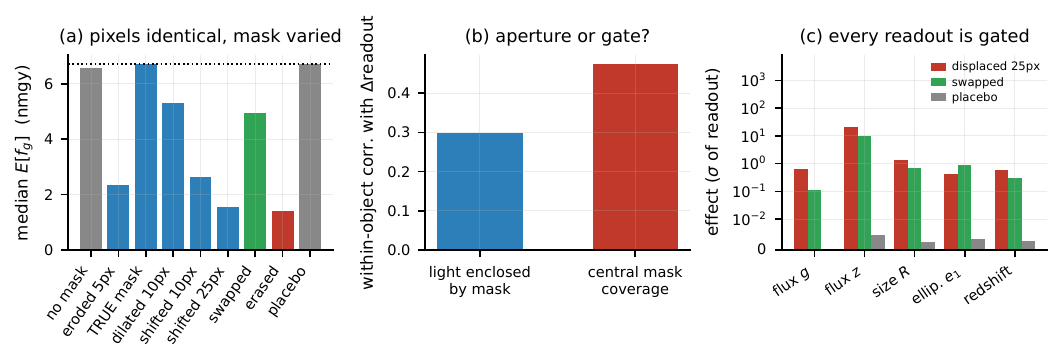}
\caption{The detection gate. \textbf{(a)} Median $g$-band flux estimate under mask
manipulations with the image pixels held byte-identical; green is another galaxy's genuine mask
(fully on-manifold), red is an erased mask (off-manifold, discounted), grey are controls.
\textbf{(b)} Mechanism discriminator: within object and across nine intervention arms, the
change in prediction tracks central mask coverage better than the light the mask encloses.
\textbf{(c)} Every readout is gated, at 110--4400 times the placebo.}
\label{fig:gate}
\end{figure}

We hold the image tokens fixed and alter only the segmentation map. Effects are expressed in
robust $\sigma$ of each readout's own spread so that flux, size, ellipticity and redshift are
comparable; the placebo displaces the right-ascension token by 250 codebook steps
($n=480$, imaging input only). Table~\ref{tab:gate} and Fig.~\ref{fig:gate} give the result.

\begin{table}[t]
\centering\small
\caption{Effect of altering only the segmentation map, in $\sigma$ of each readout's own
spread. Image pixels are byte-identical across all columns. $n=480$, imaging input.}
\label{tab:gate}
\begin{tabular}{lccc}
\toprule
Readout & Mask displaced \SI{25}{px} & Mask swapped & Placebo \\
\midrule
flux $g$            & 0.630  & 0.110 & 0.0004 \\
flux $z$            & 20.55  & 9.83  & 0.0047 \\
size $R$            & 1.313  & 0.718 & 0.0025 \\
ellipticity $e_1$   & 0.435  & 0.931 & 0.0036 \\
redshift            & 0.580  & 0.312 & 0.0028 \\
\bottomrule
\end{tabular}
\end{table}

Every readout is gated. For four of the five readouts a displaced mask is more damaging than a
swapped one---wrong \emph{location} costs more than wrong \emph{shape}. Ellipticity is the
exception, and the ordering reverses there ($0.435$ against $0.931\,\sigma$); the reversal
persists when spectra are added ($0.342$ against $0.895$), where the other four readouts still
follow the rule. In absolute terms ($n=240$) the median
$g$-band flux estimate falls from 6.71 to \SI{1.55}{nmgy} when the mask is shifted
\SI{25}{px}---the median per-object change is 79\,\%, with the pixels untouched---and to \SI{4.93}{nmgy} when another
galaxy's genuine mask is substituted.

Independent evidence suppresses the gate selectively. Adding DESI spectra reduces the redshift
effect a hundredfold ($0.312\rightarrow0.003\,\sigma$) and the flux-$z$ effect threefold, while
leaving ellipticity untouched ($0.931\rightarrow0.895$)---the pattern expected if the model
weighs evidence and the gate survives only where no counter-evidence exists.

\subsection{Mechanism: detection gating, not aperture photometry}

Across nine intervention arms the within-object fractional change in prediction correlates with
\emph{central mask coverage} at $r=0.474$ and with the \emph{light enclosed by the mask} at
$r=0.298$. Eroding the mask by \SI{5}{px} collapses the estimate while dilating it by
\SI{10}{px} barely does---the reverse of aperture behaviour and the signature of a presence
gate. Every collapsing arm converges on the same low floor.

We flag a statistic that misleads. Pooled over the graded-mask arms, the correlation between
prediction and enclosed light is 0.38---four times the within-object value---and is open to
reading as aperture photometry. It is object-to-object brightness: within the mask-swap arm,
where the mask belongs to another galaxy entirely, the across-object correlation is higher still
at 0.73, while the within-object value is only 0.09.

\subsection{Presence, not partition: evidence from real blends}

The strongest test uses no synthetic geometry. In a blended cutout the pipeline has already
partitioned the light into segments, and the catalogue flux refers to the central object alone.
We therefore compare two masks that are both genuine pipeline output---the central segment alone
versus all segments---which differ by exactly the deblending decision. Blending is expected to be among the leading
systematics for LSST-era imaging \citep{melchior2021blending,sanchez2021blending},
and modern deblenders exist precisely to control it \citep{melchior2018scarlet}.

\begin{figure}[t]
\centering
\includegraphics[width=0.72\textwidth]{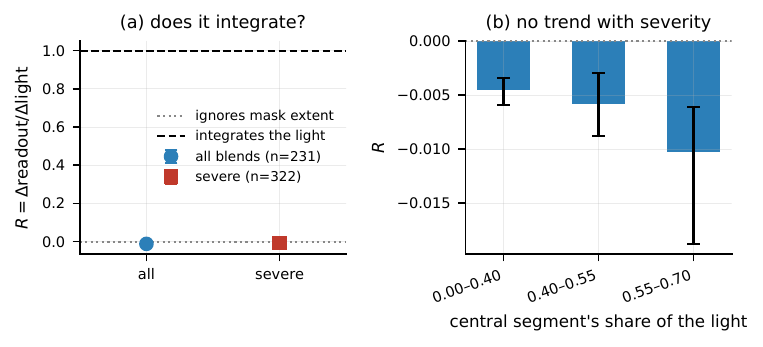}
\caption{Real blends, no synthetic geometry: only the choice of which \emph{pipeline} segments
the mask includes. \textbf{(a)} $R=\Delta$readout$/\Delta$light is zero, not one: the model
ignores how the pipeline partitioned the light. \textbf{(b)} The null holds across blend
severity, including systems where the central segment holds under 40\,\% of the light.}
\label{fig:blends}
\end{figure}

If the model integrated the light assigned to it, $R=\Delta(\text{prediction})/\Delta(\text{enclosed
light})$ would be $\approx1$. Measured on the 231 of 240 real blends with more than
\SI{0.05}{nmgy} of added neighbour light, a median \SI{1.053}{nmgy} of genuine
neighbour light is added and the readout moves by \SI{-0.016}{nmgy}: $R=-0.012$
$[-0.014,-0.008]$. Selecting for severity (322 systems, central segment $\le70\,\%$ of the
light, minimum 1.2\,\%) gives $R=-0.0064$ $[-0.0075,-0.0046]$, with no trend across severity
bins (Fig.~\ref{fig:blends}). The two masks differ in 28--39\,\% of segmentation tokens and no
object tokenises identically, so the codec resolves the deblending decision the model ignores.

\textbf{Deblending errors therefore do not propagate through this channel.} The vulnerability is
to detection and position.

\subsection{Is the gate carried by a single direction?}

If the model represents ``a source is present here'' along one axis of its internal state, then
adding that axis back should repair a readout that a displaced mask has collapsed. We test this
directly. Taking the average internal state of 256 objects under a true mask, subtracting their
average under a displaced mask, and adding the difference back at encoder block 6 recovers
46\,\% of the flux collapse at $\alpha=4$; a random vector of identical length recovers 5\,\%.
So a substantial part of the gate is carried by one direction, and it is a direction that a
subtraction of averages finds.

It is not the whole story. The per-object directions point the same way only to a consistency of
0.80, and 22 of the 768 dimensions are needed to capture 90\,\% of what remains. Two cautions on
reading these numbers. Large interventions restore the readout non-specifically, so peak recovery
means little: at $\alpha=8$ the gate direction reaches 89\,\% but the random control reaches
66\,\%, which is why we quote the pre-registered $\alpha=4$ throughout. And we inject at block 6
only, so this locates nothing about \emph{where} the gate is computed. The comparable figures in
\S\ref{sec:sae} come from a separate harness and are not on the same $\alpha$ scale.

\subsection{The behaviour generalises across channels}
\label{sec:general}

\aion{} is the only public astronomical foundation model with a detection-map input, so a
cross-model test of the same channel is not currently possible. We therefore test across
channels within the model (Table~\ref{tab:meta}).

\begin{table}[t]
\centering\small
\caption{Metadata preference. Pixels identical; the target band ($g$) is never supplied.
$n=480$.}
\label{tab:meta}
\begin{tabular}{lcc}
\toprule
Input configuration & $\dmag$ vs catalogue & Shift vs clean arm \\
\midrule
imaging only                        & 0.0665 & --- \\
imaging + true Tractor $r,i,z$      & 0.0339 & --- \\
\textbf{imaging + wrong $r,i,z$}    & \textbf{0.6210} & \textbf{0.6121} \\
imaging + true segmap               & 0.0693 & --- \\
imaging + wrong segmap              & 0.1682 & 0.1322 \\
placebo (RA moved 250 codes)        & 0.0350 & 0.0005 \\
\bottomrule
\end{tabular}
\end{table}

Corrupting the catalogue scalars moves the readout $4.6\times$ more than corrupting the
segmentation map. Some response is legitimate---galaxy colours are correlated, so $r,i,z$
genuinely inform $g$. The pathology is the comparison against ignoring the channel:
contradicted metadata leaves the model \emph{nine times worse than having no metadata at all}
(0.62 versus \SI{0.067}{mag}). A model weighing evidence would fall back toward the pixels. The
claim is therefore about the design pattern---supplying redundant, highly predictive catalogue
channels alongside raw data---and the detection gate is a comparatively mild instance of it.

\subsection{Scaling makes it worse}

\begin{table}[H]
\centering\small
\caption{The gate against model scale, $n=240$ identical objects through both checkpoints.}
\label{tab:scale}
\begin{tabular}{lccc}
\toprule
Vulnerability (mask swapped) & \texttt{aion-base} (\num{314}\,M) & \texttt{aion-large} (\num{860}\,M) & Ratio \\
\midrule
flux $g$  & 0.147 $[0.127,0.166]$ mag & 0.237 $[0.170,0.267]$ mag & 1.62 \\
redshift  & 0.0327 $[0.0266,0.0361]$  & 0.0441 $[0.0377,0.0518]$  & 1.35 \\
\bottomrule
\end{tabular}
\end{table}

The larger model defers \emph{more}. On present evidence scaling does not mitigate this and
appears to aggravate it. That scaling alone does not close a domain gap for galaxy images is
consistent with \citet{walmsley2024scaling}; that a failure mode \emph{grows} with parameter
count at fixed data is, to our knowledge, not previously reported.

\section{Incidence and cosmological consequence}
\label{sec:impact}

\subsection{How often the failure occurs}

Measured from the survey's own segmentation maps over 5000 cutouts:
\begin{itemize}\itemsep1pt
\item \textbf{3.68\,\% of targets have no segment covering their position}---an unambiguous
pipeline failure, and precisely the configuration that collapses the readout;
\item central-segment centroid offsets have median \SI{1.58}{px}, p90 \SI{12.3}{px}, p99
\SI{35.1}{px}, with 11.7\,\% beyond \SI{10}{px}.
\end{itemize}

\subsection{Impact on tomographic mean redshift}

Photometric-redshift requirements constrain the bin mean: the LSST DESC Science Requirements
Document \citep{lsstdesc2018srd} requires $\dzreq<0.003$ for the Y10 large-scale-structure
sample; we adopt that requirement because our bins are low-redshift BGS-like galaxies, and note
that the Y10 weak-lensing requirement is three times tighter, against which the exceedances
below are correspondingly larger. See also
\citet{ivezic2019lsst,lsstscience2009sciencebook} for the survey, and
\citet{schmidt2020evaluation,benitez2000bayesian,salvato2019flavours,myles2021redshift} for
photometric-redshift methodology and calibration. We separate two scenarios because they rest on
different evidence.

\begin{table}[t]
\centering\small
\caption{Induced tomographic bias $\dzreq$, against the requirement 0.003. A detection miss is
represented by a segmentation field transplanted from the 184 measured misses, not by an
all-zero field. Redshifts read with the posterior median. Bold exceeds the requirement. $n=960$.
Per-bin values are the single realisation plotted in Fig.~\ref{fig:impact}b; the Scenario-1
ensemble over 40 miss assignments has worst-bin median $0.71\times$ the requirement.}
\label{tab:tomo}
\begin{tabular}{lcccc}
\toprule
& \multicolumn{2}{c}{Scenario 1: measured miss rate only} & \multicolumn{2}{c}{Scenario 2: + displacements} \\
\cmidrule(lr){2-3}\cmidrule(lr){4-5}
Bin & Imaging only & + spectra & Imaging only & + spectra \\
\midrule
$[0.00,0.15)$ & 0.00126 & 0.00000 & \textbf{0.02479} & 0.00111 \\
$[0.15,0.25)$ & 0.00204 & 0.00002 & 0.00151 & 0.00000 \\
$[0.25,0.35)$ & 0.00047 & 0.00000 & \textbf{0.00409} & 0.00008 \\
$[0.35,1.00)$ & 0.00024 & 0.00000 & \textbf{0.01797} & 0.00004 \\
\midrule
worst / requirement & $0.68\times$ & $0.006\times$ & $8.3\times$ & $0.37\times$ \\
\bottomrule
\end{tabular}
\end{table}

Scenario 1 applies only the measured 3.68\,\% miss rate. How a miss is \emph{represented}
turns out to decide the answer, so we state it explicitly. The obvious choice---zeroing the
segmentation field---is wrong: it is the \emph{erased} arm of \S\ref{sec:gate}, the most
damaging single intervention we measure, and it does not occur in the data. Of the 184 misses
in our 5000-cutout scan, \emph{none} returns a blank field; the pipeline fails to place a
segment on the target while still labelling its surroundings, at a median central coverage of
0.146 against 0.788 for detected objects. We therefore represent a miss by transplanting a
segmentation field drawn from those 184 measured cases, holding the object's own pixels fixed.
Scenario 2 additionally displaces the remaining masks by draws from the observed
segment-centroid offset distribution; because that distribution mixes genuine astrometric and
detection error with real morphological asymmetry, it is a plausible magnitude rather than a
validated error rate.

\paragraph{Construction and assumptions.} Because a reader cannot audit this step from the
figures alone, we set out how the number is built and what it rests on. (i) The bias is
computed on $n=960$ galaxies drawn at random from the DESI BGS-bright cross-match, binned into
four tomographic bins by \emph{true} spectroscopic redshift, not by the model's own estimate;
binning on the estimate would mix in selection effects we are not trying to measure, and would
change the numbers. (ii) The quantity is the shift in each bin's mean redshift between a run
with the survey's true masks and a run with perturbed masks on the \emph{same} objects, so
per-object scatter cancels and only the coherent component survives---which is what the
requirement constrains. (iii) The $3.68\,\%$ rate is directly counted from 5000 real cutouts and
is not modelled. (iv) In Scenario 2 the two perturbations act on disjoint sets---missed objects
receive a transplanted miss field, the remaining objects receive a positional displacement---so
the $8.3\times$ does not double-count the miss. (v) The displacement draws come from the observed
distribution of central-segment centroid offsets, which mixes genuine astrometric and detection
error with real morphological asymmetry; treating all of it as error is what makes Scenario 2 a
plausible magnitude, and an upper bound on the positional contribution. (vi) Which objects are missed is itself a draw, so
Scenario 1 is reported as an ensemble of 40 assignments rather than one realisation; Scenario 2
is a single realisation and is quoted as an upper bound only. Taken together, Scenario 1 is a
bounded estimate under the assumptions above and Scenario 2 an upper bound; measuring the
systematic as realised in a survey would require the injection frameworks noted in the
limitations.

\begin{figure}[t]
\centering
\includegraphics[width=\textwidth]{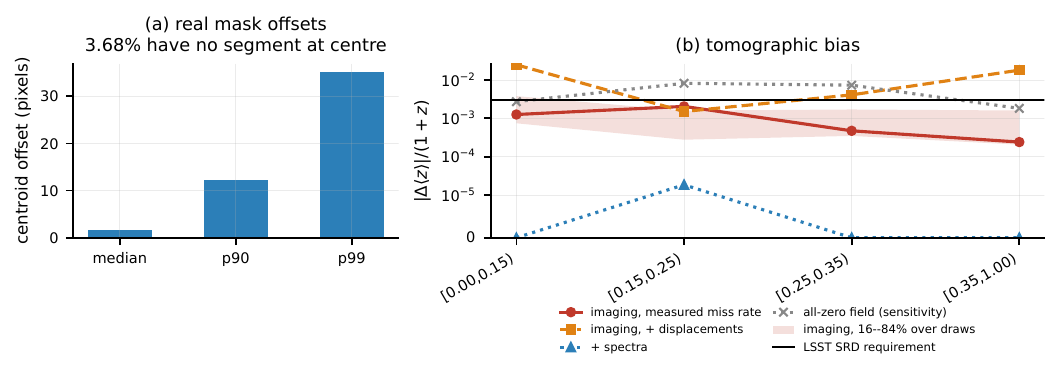}
\caption{Incidence and consequence. \textbf{(a)} Distribution of central-segment centroid
offsets measured from 5000 real Legacy Survey segmentation maps. \textbf{(b)} Induced bias in
tomographic mean redshift against the LSST DESC requirement. The measured detection-miss rate
alone reaches two-thirds of the requirement without breaching it; adding positional errors of
the observed magnitude breaches it in three of four bins, and adding spectra removes the effect.
The shaded band is the 16--84\,\% range of the imaging, measured-miss-rate curve
over 40 miss assignments; the solid curve is one realisation from within it. The grey curve
repeats the imaging-only propagation with an all-zero segmentation field instead of an
empirically sampled one: the sensitivity quantified in \S\ref{sec:impact}.}
\label{fig:impact}
\end{figure}

The result is sensitive to this choice by a factor of four, which is why we state it
explicitly: the same miss rate propagated through an all-zero field would breach the requirement
by $2.7\times$ in two of four bins, against $0.88\times$ for the empirical representation
(Fig.~\ref{fig:impact}b).

Across 40 miss assignments the worst-bin bias has median $0.71\times$ the requirement with a
16--84\,\% range of $0.47$--$1.27\times$, and 12 of the 40 realisations exceed the requirement in
at least one bin. The scatter matters as much as the median: a single routine pipeline failure
consumes roughly two-thirds of the allowance for the mean redshift of a tomographic bin in the
typical case, and the whole of it in about a third of realisations, on a model whose output
carries no sign that anything has gone wrong. In the median \aion{} stays inside the
requirement, but not with margin to spare.

What does breach the requirement is position. Adding displacements of the observed magnitude
takes the worst bin to $8.3\times$ and puts three of four bins over, which is the same ordering
\S\ref{sec:gate} found under controlled intervention---wrong location costs more than wrong
shape, and \S\ref{sec:gate} showed deblending errors do not propagate at all. The risk this
model carries into a cosmological analysis is astrometric and detection-positional, not
photometric. Spectroscopy removes all of it. The systematic is thus specific to photometric-only
inference, and it is invisible in the model's output.

\subsection{Are the misses a random subsample?}
\label{sec:corrmiss}

Scenario 1 draws the missed objects uniformly, and real detection misses are not uniform. We
measured the selection instead of assuming it. Over the same 5000 cutouts, the miss probability
rises monotonically with $r$-band magnitude from 0.33\,\% in the brightest eighth of the sample
to 7.62\,\% in the second-faintest---a factor of 20, with missed objects a median
\SI{0.49}{mag} fainter than detected ones (Mann--Whitney $p=5\times10^{-24}$;
Appendix~\ref{app:miss} gives the bin table). The selection is real and it is strong.

Re-drawing the same number of misses with probability proportional to that measured
$p(\text{miss}\mid\text{mag})$ does shift which objects are lost: the median redshift of the
missed population moves from 0.240 to 0.274 across 40 paired draws ($p=1\times10^{-9}$). The
tomographic bias, however, barely responds. The worst-bin median rises from $0.71\times$ to
$0.88\times$ the requirement, a ratio of 1.24, but the correlated arm is higher in only 23 of 40
paired draws and the difference sits well inside the realisation scatter (Wilcoxon $p=0.50$; a
lower-variance statistic, the mean across bins, gives 1.15 at $p=0.40$). Realisations breaching
the requirement go from 12/40 to 17/40 ($p=0.18$). Supplying spectra leaves the bias at
$0.007$--$0.009\times$ the requirement under either rule.

Magnitude selection therefore fails to propagate into the bin means, for a reason visible in the
sample: in a magnitude-limited bright-galaxy survey at these redshifts, apparent magnitude and
redshift are only loosely coupled, so a strong cut in magnitude is a weak one in $z$. The
uniform draw is wrong in detail and immaterial in effect at this precision. We would not expect
that to hold for a deeper, higher-redshift sample where the two are tightly coupled.

\subsection{Mitigation}

\begin{table}[H]
\centering\small
\caption{Cost of withholding the detection channel, against the vulnerability it removes.
$n=480$. Bracketed values are bootstrap 16--84\,\% intervals; flux $g$ intervals are shown in
Fig.~\ref{fig:mitig}a. Zero entries in the last column are not measured---no swap arm is run for
configurations without a real mask.}
\label{tab:mitig}
\begin{tabular}{lccc}
\toprule
Configuration & flux $g$ ($\dmag$) & redshift $|\Delta z|$ & Swap vulnerability \\
\midrule
imaging only            & 0.0665 & 0.1169 $[0.1143,0.1200]$ & \textbf{0} \\
imaging + real mask     & 0.0693 & 0.1273 $[0.1253,0.1314]$ & 0.132 mag / 0.031 \\
imaging + generic disc  & 0.2215 & 0.1566 $[0.1510,0.1599]$ & 0 \\
+ spectra, no mask      & 0.0534 & 0.0034 $[0.0033,0.0035]$ & \textbf{0} \\
+ spectra + real mask   & 0.0575 & 0.0038 $[0.0036,0.0039]$ & 0.086 mag / 0.0004 \\
\bottomrule
\end{tabular}
\end{table}

Withholding the mask is free: redshift improves with non-overlapping intervals in both bases,
and flux $g$ is unchanged within the intervals (an independent sample gave \SI{0.005}{mag} the
other way, so we describe it as neutral). It removes the vulnerability entirely.

\begin{figure}[t]
\centering
\includegraphics[width=0.82\textwidth]{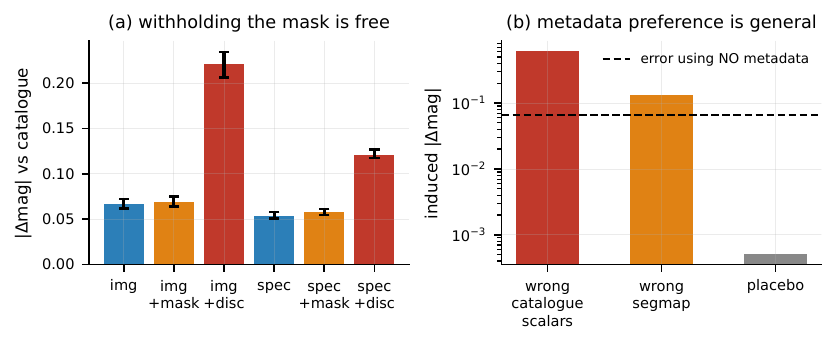}
\caption{\textbf{(a)} Withholding the detection mask costs nothing measurable, while a generic
substitute mask is far worse than either option. \textbf{(b)} The preference is general:
corrupted catalogue photometry moves the readout further than a corrupted segmentation map, and
both leave the model worse than the dashed line, which is the error using no metadata at all.}
\label{fig:mitig}
\end{figure}

A generic centred disc is much worse than either option ($+220\,\%$ on flux $g$;
Fig.~\ref{fig:mitig}a), so the channel carries more than a presence flag, and
substituting a neutral mask degrades the readout instead of protecting it. The recommendation is
to withhold the channel when the detection cannot be vouched for.

\section{Tokeniser-limited behaviour}
\label{sec:tok}

\subsection{Codebook collapse}

Codebook under-utilisation is a known failure mode of vector-quantised models
\citep{yu2022vector,huh2023straightening,mentzer2023finite}. Measured directly over
$1.15\times10^{6}$ tokens from real cutouts, and splitting image tokens by whether the
$4\times4$ patch overlaps the source---a $96\times96$ cutout is mostly sky, and the raw
histogram is dominated by a single ``empty'' code---we obtain Table~\ref{tab:codes}.

\begin{table}[H]
\centering\small
\caption{Measured codebook utilisation. Perplexity is the effective number of states.}
\label{tab:codes}
\begin{tabular}{lcc}
\toprule
& Codes emitted & Effective size (perplexity) \\
\midrule
image, source patches       & 237 / 4375 & \textbf{28.2} \\
image, sky patches          & 356 / 4375 & 5.2 \\
spectrum                    & 1024 / 1024 & \textbf{934} \\
image, HSC (864\,k tokens)  & 312 / 4375 & 2.5 (all-patch; cf. Legacy 7) \\
\bottomrule
\end{tabular}
\end{table}

An embedding row-norm heuristic gives 23.2\,\% for Legacy and 9.6\,\% for HSC and overstates
both: a rarely emitted code still receives gradient and retains its norm. Every code HSC emits
is also Legacy-live, confirming the containment seen in the embedding tables. Utilisation is
unchanged between the two model scales, localising the effect to the tokeniser and the data
rather than to capacity.

\subsection{The image codes are a brightness ladder}

\begin{figure}[t]
\centering
\includegraphics[width=\textwidth]{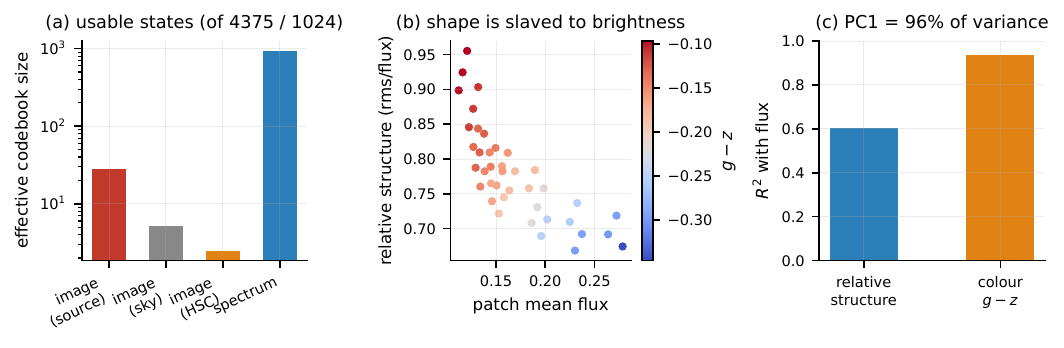}
\caption{Tokeniser limits. \textbf{(a)} Effective codebook size: the image codec resolves
$\sim$28 states on source patches against 934 for the spectrum codec. \textbf{(b)} The most-used
image codes lie on a single brightness axis. \textbf{(c)} After removing the trivial flux
scaling, relative structure and colour remain largely predicted by flux.}
\label{fig:tok}
\end{figure}

Decoding the most-used codes \emph{in context}---overwriting one central grid token in real
cutouts and differencing, since decoding a uniform grid returns the decoder's prior---the 40
codes carrying 93.3\,\% of source patches lie on a one-dimensional manifold. PC1 of (flux,
structure, colour) explains 96.3\,\% of the between-code variance with near-isotropic loadings,
the signature of a single latent scalar, and $R^2(\text{flux},\text{PC1})=0.999$ identifies it
as brightness. Removing the trivial flux scaling, relative structure (rms/flux) varies by only
8.7\,\% across all 40 codes and colour is 94\,\% predicted by flux (Fig.~\ref{fig:tok}).

Morphology can therefore only be represented in the spatial arrangement of ladder positions
across the grid, never within a token. With the codebook collapse above, this is a concrete, tokeniser-level
account of why \aion{}'s imaging results trail its spectroscopic ones, and it implies that
scaling the transformer cannot close the gap. It is a caution for morphology applications in
particular, where human-labelled vocabularies are rich
\citep{walmsley2022decals,walmsley2023gzdesi,walmsley2023zoobot}.

\subsection{Redshift is quantisation-limited}

\texttt{tok\_z} is quantised on a \emph{linear} grid over $z\in[0,6]$ with $\Delta z=0.005865$,
unlike the $\sim$31 empirical-CDF scalars, and 27\,\% of its codes are never used. With spectra,
the median redshift error is $0.497\times$ the grid spacing and 80.8\,\% of objects place
$>90\,\%$ of posterior mass in a single bucket. DESI's measured random redshift error for bright galaxies is
$\sim\SI{10}{\kilo\metre\per\second}$ \citep{lan2023visual}, i.e. $\sigma_z\approx3.3\times10^{-5}$
and some 180 times finer than the grid. The precision bottleneck
is the tokeniser.

\section{Posterior calibration and the redshift mechanism}
\label{sec:calib}

\subsection{Calibration}

\citet{parker2025aion} treat single-token outputs as categorical posteriors, confining their
calibration caveat to ``sequences of tokens longer than a single token'', and report no
calibration test of the single-token case. On \texttt{tok\_z} ($n=3000$):
\begin{itemize}\itemsep1pt
\item \textbf{The posterior mean is unusable.} With photometry alone the median $|\Delta z|$ is
0.236 for the mean against 0.050 for the median; the linear grid drags the mean upward whenever
the posterior is broad.
\item Photometry-only posteriors are close to calibrated (coverage 0.513/0.662 at nominal
0.50/0.68).
\item With spectra the tails are far too thin: the nominal 99\,\% interval covers 75\,\%. Given
\S\ref{sec:tok}, the natural reading is the quantisation floor rather than unrepresented
uncertainty.
\end{itemize}

\subsection{Redshift is line-locked, but not line-brittle}

Ablating windows of eight spectrum tokens ($\sim$\SI{205}{\angstrom}, matched to the codec's
measured $\sim$\SI{280}{\angstrom} effective resolution) and stacking the response in the rest
frame gives a contrast of 32.5, against 4.4 in the observed frame and 2.95 for a
shuffled-redshift null. The peak lies at \SI{6643}{\angstrom}---H$\alpha$, within one bin
(Fig.~\ref{fig:z}a). Causal weights: H$\alpha$+[N\,\textsc{ii}] 25.0, Mg\,b 15.4, Na\,D 12.8,
Ca\,H\&K / \SI{4000}{\angstrom} break 8.1, [O\,\textsc{iii}] 6.8. The Ca\,\textsc{ii} triplet
registers 0.10, as it must: at the median redshift it lies beyond the codec's wavelength limit.

We pre-registered the prediction that line-locking implies catastrophic, \emph{predictable}
failure. It is falsified. Ablating the H$\alpha$ window breaks 6.1\,\% of previously correct
redshifts where an identical-width quiet window breaks one of the same 2865 objects, but only 5.7\,\% of
the 174 induced failures land on line-ratio aliases against a 43.4\,\% random null, and all are
modest upward shifts. Testing the estimators directly on a 612-object subsample, the failure
rate is 5.7\,\% with the posterior mean---consistent with the 6.1\,\% above to within sampling
error---but 1.1\,\% with the median, with the median error essentially unchanged
($0.00285\rightarrow0.00293$) while posterior entropy rises $2.37\times$. \aion{} responds to
losing H$\alpha$ approximately correctly: it widens its posterior. It is line-\emph{dependent}
without being line-\emph{brittle}.

\section{Sparse dictionaries are unreliable causal handles}
\label{sec:sae}

A sparse autoencoder is a learned dictionary over the internal states of \S\ref{sec:methods}:
it fits a set of directions such that any activation can be rewritten as a sum of just $k$ of
them, in the hope that individual dictionary entries correspond to individually meaningful
features. It is the standard modern tool for this question, and the reason to want one here is
that its entries would be candidate handles on the gate---more targeted than the single
difference-in-means direction of \S\ref{sec:gate}.

Ours compresses the activations well. A BatchTopK autoencoder
\citep{bussmann2024batchtopk} with 8192 dictionary entries and $k=32$ active at a time, trained
on \num{288000} paired activations from block 6 (\num{576000} rows, pooling the true-mask and
displaced-mask conditions so the dictionary sees both), reconstructs at a fraction of variance
unexplained of 0.032, where 0 would be perfect. Principal component analysis restricted to the
same number of components---the fair like-for-like, since it is also a linear code of the same
sparsity---reaches only 0.242, so the dictionary is $7.6\times$ better as a compression. It also
contains entries that fire far more under a true mask than a displaced one: features that look,
by inspection, like the gate.

The question we pre-registered is whether looking like the gate buys control over it. Building a
steering vector from the most gate-selective entries and adding it at matched norm, exactly as
in \S\ref{sec:gate}: does it beat the plain difference in means at restoring the collapsed
readout?

Because one point in hyperparameter space cannot answer a question about a method, we test 15
dictionaries: widths 2048, 8192 and 32768 ($2.7\times$ to $43\times$ overcomplete) crossed with
$k\in\{16,32,64\}$, plus two further seeds at $k=32$ for each width. Every dictionary goes
through the identical causal test against identical baselines at identical norm
(Appendix~\ref{app:sweep}).

\begin{figure}[t]
\centering
\includegraphics[width=0.85\textwidth]{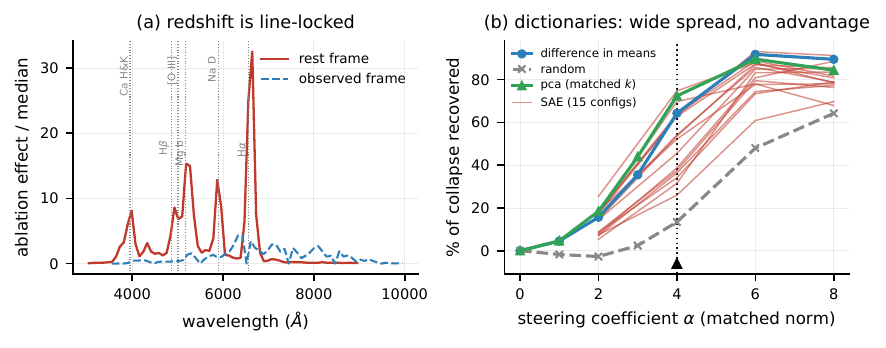}
\caption{\textbf{(a)} Rest-frame stacked ablation profile of the redshift posterior, against the
observed-frame version. The rest-frame peak at H$\alpha$ and the secondary features are absent
in the observed frame. \textbf{(b)} Causal steering at matched intervention norm, for all 15
dictionaries of the sweep (thin red) against the baselines. At the pre-registered $\alpha=4$
(marker) every dictionary beats the random control, two of fifteen beat difference-in-means, and
the spread across seeds at fixed width and sparsity reaches 18.4 points. PCA is drawn at $k=32$;
$k=16$ and 64 give 68.1 and 73.4\,\%. At $\alpha\ge6$ large-norm perturbations inflate the
readout non-specifically---the random control itself reaches 48\,\%---so peak recovery is not the
statistic.}
\label{fig:z}
\end{figure}

The answer depends on which dictionary, and that is the result. Recovery at $\alpha=4$ spans
26.0--74.7\,\% with median 52.2\,\% (Fig.~\ref{fig:z}b). All 15 beat the matched-norm random
control (13.4\,\%), so the dictionary direction always carries real signal. But only two beat
difference-in-means (64.3\,\%), and only one---32768 features at $k=16$, recovering
74.7\,\%---beats PCA at its own matched sparsity, which reaches 68.1--73.4\,\% across $k$ and is
the strongest method we test. Thirteen of fifteen dictionaries lose to a baseline that costs one
subtraction.

The spread is not hyperparameter sensitivity that careful tuning would remove. Holding width and
sparsity fixed and varying only the random seed moves recovery by up to 18.4 percentage points
(62.8/63.9/45.5\,\% at width 2048, $k=32$), which is larger than the gap between the median
dictionary and difference-in-means. The dictionary arm is by a wide margin the noisiest
measurement in this paper, and its variance is not reducible by anything an author controls
except averaging over seeds.

Nor can the good dictionaries be identified in advance from reconstruction quality, which is the
quantity dictionary learning optimises. Across the sweep the two are decoupled: the best
reconstructor (FVU 0.023) recovers 53.5\,\%, while the best steerer reconstructs \emph{worse}
than average (FVU 0.067) with only 10.5\,\% of its features alive. A practitioner choosing a
dictionary by FVU---the standard criterion---has no purchase on which one will steer.

Two consequences follow. For this target, sparse dictionaries do not earn their cost: they are
beaten on average by difference-in-means and on average and at best by PCA, at a fraction of the
compute. More generally, a causal evaluation of a dictionary that reports a single seed is
uninformative about dictionaries, because the seed-to-seed spread exceeds the effect being
claimed. Reports on both sides of this question in scientific models
\citep{rosenfeld2026sparse,marks2025sparse,zhang2024towards,wu2025euclid} are, on the evidence
here, sampling a distribution wide enough to contain both answers.

\section{Discussion}
\label{sec:disc}

The unifying description of \aion{}'s behaviour is a \emph{metadata preference}: given a
catalogue-derived channel that was highly predictive during training, the model relies on it
rather than on the raw signal, and does not discount it when the two conflict. The detection
gate is the most consequential instance because detection failures are common (3.68\,\%) and
because the resulting bias is coherent across objects, which is what makes it a cosmological
systematic instead of added noise. Its magnitude, however, depends on how a failure is
represented, and \S\ref{sec:impact} shows that the intuitive representation overstates it
fourfold---a caution that applies to any attempt to price a learned model's failure modes by
intervening on its inputs.

This is a property of the design pattern, not a coding error. Supplying redundant catalogue
products alongside raw data gives the model an easier route to the answer during training, and
nothing in the masked-modelling objective penalises over-reliance on it. Two observations
sharpen the concern: the effect grows with model scale, and it is invisible in the output---a
wrong mask produces a confident wrong answer, not a widened posterior. The contrast with
\S\ref{sec:calib} is instructive: removing genuine spectral information \emph{does} widen the
posterior appropriately. The model handles missing information well and contradictory metadata
badly.

The tokeniser results point the same way for a different reason. An image tokeniser with 28
effective states arranged on a brightness ladder cannot represent morphology within a token, and
utilisation does not improve with scale. Practitioners comparing \aion{}'s imaging and
spectroscopic performance should attribute the gap to the codec before the transformer.

More broadly, this is a case where interpretability produced an actionable systematic rather
than an explanation. Here the default modern tool, a sparse dictionary, was
beaten by far simpler linear methods on a causal task in thirteen of fifteen configurations, and
varied more with its random seed than with either baseline's margin (\S\ref{sec:sae}). That sits
alongside broader reports that sparse-dictionary features in scientific models are only
intermittently consistent and do not map cleanly onto standard physical decompositions
\citep{rosenfeld2026sparse}, though on non-causal alignment metrics sparse dictionaries can
outperform PCA \citep{wu2025euclid}. Parallel efforts are
under way in other scientific domains---fluid dynamics \citep{hu2026cfd}, quantum many-body
systems \citep{qi2026quantum} and genomics \citep{guan2025genelm}---so the question of which
tools transfer is a general one. That astronomy supplies
ground-truth physical labels and forward models---an advantage language interpretability lacks
\citep{cranmer2020symbolic,wetzel2025physics,lieu2025astronomy}---makes it a good testbed for
deciding which interpretability tools actually earn their cost.

\paragraph{Limitations.} All results are for one model family on DESI BGS-bright galaxies at
Legacy depth; the miss rate, and the blending null in particular, may differ in deeper or more
crowded imaging. Source-injection frameworks \citep{suchyta2016balrog,everett2022balrog} would
give a controlled test of observing-condition dependence and were not used here. The scaling
evidence is two points. The corrupted-scalar arm of \S\ref{sec:general} bounds the behaviour
rather than estimating a field rate. The miss representation of \S\ref{sec:impact} is resampled
from only 184 measured cases in one survey at one depth, so its shape---not the 3.68\,\% rate,
which is directly counted---is the least well constrained input to that number. The
magnitude-selection test of \S\ref{sec:corrmiss} conditions on apparent magnitude alone;
surface brightness, size and local blending are plausible additional predictors of a miss that
we have not separated, and the null it reports is specific to this sample's weak
magnitude--redshift coupling. Tomographic bins in \S\ref{sec:impact} are defined on true
redshift rather than on the model's own estimate, isolating the mask effect from
binning-induced selection. The dictionary sweep varies width, sparsity and seed but stays at
one layer, with features chosen by activation gap rather than by measured causal effect; and
dictionary training on this hardware is not bit-reproducible, so the seed spread we quote
includes run-to-run nondeterminism as well as seed variance. Finally, three
judgements this paper rests on are measurable but not settled by our analysis: whether a
3.68\,\% detection-miss rate is typical of other surveys and depths, whether DESI BGS-bright is
representative for the claims made, and whether \SI{0.13}{mag} is material for the science cases
a given reader cares about. Each is a question about external validity, and each would be
settled by repeating this audit on another survey.

\section{Conclusions}

\begin{enumerate}\itemsep2pt
\item \aion{} defers to catalogue-derived metadata over pixels across every readout, and does
not discount metadata that contradicts the image.
\item The mechanism is detection gating---presence at the field centre---not aperture
photometry; deblending errors do not propagate, but detection and positional errors do.
\item The effect grows with model scale.
\item In photometric-only inference, the measured detection-miss rate alone consumes a median
two-thirds of the LSST DESC error budget for tomographic mean redshift and exceeds it in 12 of
40 realisations; observed positional errors take the worst bin to $8.3\times$ the requirement.
Drawing the misses by their measured magnitude dependence rather than uniformly does not change
this.
\item Withholding the detection channel removes it at no measurable cost; a generic substitute
mask is worse than either option.
\item The image tokeniser is collapsed to $\sim$28 effective brightness-ordered states, and the
redshift readout is quantisation-limited; both persist under scaling.
\item Sparse dictionaries reconstruct far better than PCA yet are unreliable as causal
handles: across 15 dictionaries recovery spans 26--75\,\%, moves by up to 18 points on the seed
alone, and is not predicted by reconstruction quality. Thirteen of fifteen lose to
difference-in-means.
\end{enumerate}

\paragraph{Practical recommendations.} Withhold the detection channel when it cannot be vouched
for; read the posterior median, never the mean; do not expect spectroscopic redshift precision;
query one target modality per forward pass (Appendix~\ref{app:bugs}); and attach probes to the
encoder output rather than to \texttt{model.encode()}.

\appendix

\section{Engineering notes on the released code}
\label{app:bugs}

These are recorded so that others reproducing this work do not lose time to them, and because
two of them silently change results rather than raising an error. All five surfaced from using the model in ways its authors
had no need to, and none affects the results reported in the \aion{} paper. All observations are against the PyPI package
\texttt{polymathic-aion} 0.0.2 with weights \texttt{aion-base} at revision
\texttt{40541618} and \texttt{aion-large} at \texttt{cfdb89e8}; we have not verified them
against later revisions, and items 1 and 3 in particular may already be fixed upstream.

\begin{enumerate}\itemsep2pt
\item \textbf{Batched decoder targets corrupt every readout.}
\texttt{AION.forward(\ldots,\,target\_modality=[A,B,C])} supplies an all-zero
\texttt{decoder\_attention\_mask} which the adapter converts to ``all positions masked''.
Harmless for one decoder token; with more, decoder self-attention becomes uniform. Measured on
redshift with image+spectrum input: median $|\Delta z| = 0.0038$ with \texttt{[Z]}, 0.833 with
\texttt{[FluxG, Z]}, 2.350 with \texttt{[FluxG, FluxZ, Z]}. Use one target per call.
\item \textbf{Version skew.} Repository HEAD adds spectrum sentinel padding absent from released
0.0.2, and the two produce different spectrum tokens. Real DESI spectra require a trailing
$\lambda=99999$ point; omitting it changes 20\,\% of the 273 tokens while remaining invisible in
end-to-end accuracy.
\item \textbf{Specification mismatch in the documentation.} The published image-FSQ
specification ($\{8,5,5,5\}$, $\sim2^{12}$ codes) matches the segmentation codec rather than the
image codec, which we measure as $7\cdot5^4=4375$ with embedding dimension 5. Anyone sizing a
codebook analysis from the documented figure will be wrong by \num{-289} codes.
\item \textbf{HSC image decoding raises} an attribute-name error on the path we exercised;
tokenisation is unaffected, so this does not touch any HSC result reported here.
\item Preprocessing mutates caller tensors in place, and \texttt{tok\_z} carries a 1025th
sentinel code with no value bucket.
\end{enumerate}

\section{The measured detection-miss selection}
\label{app:miss}

Table~\ref{tab:missmag} is the empirical $p(\text{miss}\mid\text{mag})$ underlying
\S\ref{sec:corrmiss}, counted over the same 5000-cutout scan that gives the 3.68\,\% rate. Bins
are octiles of $r$-band magnitude, so each holds roughly 605 objects. Rates are raw counts; the
weights used for the correlated draw shrink these toward the global rate with 25 pseudo-counts,
so that a bin which happens to contain no misses does not make its objects unmissable by
construction.

\begin{table}[H]
\centering\small
\caption{Detection-miss rate against $r$-band magnitude, over 5000 real cutouts (184 misses).
The trend is monotone across seven of eight bins; missed objects are a median \SI{0.49}{mag}
fainter than detected ones, at Mann--Whitney $p=5\times10^{-24}$.}
\label{tab:missmag}
\begin{tabular}{lrr}
\toprule
$r$ magnitude & $n$ & $p(\text{miss})$ (\%) \\
\midrule
$13.37$--$18.05$ & 605 & 0.33 \\
$18.05$--$18.67$ & 604 & 0.83 \\
$18.67$--$19.06$ & 605 & 0.99 \\
$19.06$--$19.34$ & 604 & 1.99 \\
$19.34$--$19.55$ & 604 & 3.81 \\
$19.55$--$19.81$ & 605 & 6.12 \\
$19.81$--$20.03$ & 604 & 7.62 \\
$20.03$--$20.30$ & 605 & 6.61 \\
\bottomrule
\end{tabular}
\end{table}

\section{The dictionary sweep in full}
\label{app:sweep}

Every configuration in Table~\ref{tab:sweep} was trained on the same pooled activation matrix
(\num{576000} rows spanning the gated and ungated conditions), evaluated on the same 192 held-out
objects, and steered at the same intervention norm $\|v\|=\|v_{\text{diff-in-means}}\|$, so the
only quantities varying down the table are dictionary width, sparsity and seed. PCA is refit at
each $k$ so that ``matched sparsity'' remains true. The baselines at $\alpha=4$ are
difference-in-means 64.3\,\%, random 13.4\,\%, and PCA 68.1/72.4/73.4\,\% at $k=16/32/64$.

\begin{table}[H]
\centering\small
\caption{All 15 dictionaries. Recovery is the fraction of the gate-induced flux collapse restored
at $\alpha=4$; bold exceeds difference-in-means (64.3\,\%). The first nine rows are the
width-by-$k$ factorial at seed 0; the last six vary the seed at $k=32$.}
\label{tab:sweep}
\begin{tabular}{rrrccrr}
\toprule
Width & $k$ & Seed & FVU & FVU (PCA, same $k$) & Alive (\%) & Recovery (\%) \\
\midrule
2048 & 16 & 0 & 0.065 & 0.350 & 93.7 & 38.7 \\
2048 & 32 & 0 & 0.049 & 0.242 & 99.9 & 62.8 \\
2048 & 64 & 0 & 0.039 & 0.149 & 100.0 & 63.0 \\
8192 & 16 & 0 & 0.052 & 0.350 & 51.6 & 37.4 \\
8192 & 32 & 0 & 0.032 & 0.242 & 89.0 & 31.5 \\
8192 & 64 & 0 & 0.023 & 0.149 & 99.1 & 53.5 \\
32768 & 16 & 0 & 0.067 & 0.350 & 10.5 & \textbf{74.7} \\
32768 & 32 & 0 & 0.048 & 0.242 & 26.6 & 54.2 \\
32768 & 64 & 0 & 0.056 & 0.149 & 52.4 & 36.4 \\
2048 & 32 & 1 & 0.049 & 0.242 & 99.9 & 63.9 \\
2048 & 32 & 2 & 0.049 & 0.242 & 99.9 & 45.5 \\
8192 & 32 & 1 & 0.032 & 0.242 & 89.4 & 26.0 \\
8192 & 32 & 2 & 0.031 & 0.242 & 90.1 & 34.2 \\
32768 & 32 & 1 & 0.048 & 0.242 & 28.9 & \textbf{69.7} \\
32768 & 32 & 2 & 0.051 & 0.242 & 28.4 & 52.2 \\
\bottomrule
\end{tabular}
\end{table}

\section*{Data and code availability}

All code, per-experiment JSON and the figure-generation script are available at
\url{https://github.com/Kendiukhov/astro-mechinterp}. The
repository contains \texttt{astron\_mi/} (model loading, tokenisation, activation hooks, physics
decoding, dictionary learning), 23 experiment scripts in \texttt{experiments/}, the recorded
outputs in \texttt{results/}, and the \LaTeX{} source of this manuscript.

Two files are the entry points for verification. \texttt{docs/claims-to-code.md} maps every
numbered claim in this paper to the script that produced it and the JSON key path that holds it.
\texttt{RESULTS.md} is the full experiment log, including measurements not reported here and the
caveats attached to each. Every figure is regenerated from the recorded JSON by
\texttt{experiments/make\_paper\_figures.py} with no live model call, so \texttt{make figures}
reproduces all six figures in minutes without the data or the weights; a test suite checks that
each cited result file is present and that the figure pipeline runs.

Reproducing the JSON itself requires the data and a GPU-class device. All measurements are
against \texttt{polymathic-aion} 0.0.2 with weights \texttt{aion-base} revision
\texttt{40541618} and \texttt{aion-large} revision \texttt{cfdb89e8}; \texttt{requirements.txt}
pins the environment. The data are public and not redistributed: \texttt{docs/data.md} documents
the expected on-disk schema and how to rebuild the \SI{115}{GiB} PROVABGS--DESI--Legacy
cross-match \citep{hahn2023provabgs,dey2019legacy,desi2024edr} and the MultimodalUniverse HSC
PDR3 shard \citep{mmu2024multimodal,aihara2022hscpdr3}.

\section*{Acknowledgements}

This audit was only possible because the \aion{} authors released weights, tokenisers and
training code openly, including the components whose limitations we report here. We thank the
Legacy Survey, DESI and Hyper Suprime-Cam collaborations for the public data products this work
depends on.

\paragraph{Use of generative AI.} A large language model (Anthropic Claude) was used
substantially throughout this work: to write the analysis library and the experiment scripts, to
run the experiments, to generate the figures, and to draft this manuscript. It was used as an
instrument under the author's direction; the author is responsible for all content.

Because generative AI was used for calculation, analysis and data visualisation, we set out the
steps taken to validate it. (i) Every numeric claim in this paper is checked programmatically
against the recorded JSON that produced it, and the checking script is in the repository; the
figures are regenerated from that same JSON with no model call, so a figure cannot disagree with
the text. (ii) The manuscript was audited against the stored results by an independent
adversarial pass, which found and corrected defects including a stale table value that had
propagated into the abstract, a correlation quoted from a code comment rather than from output,
and a negative control reported as zero when it was one. (iii) Experiments were checked for
protocol fidelity against the experiment they were designed to extend; one dictionary-learning
sweep was discarded and re-run because it had trained on the wrong activation set, and the
superseded run is retained in the repository. (iv) All \num{84} bibliography entries were
machine-checked to resolve to a live DOI, arXiv identifier or archival record. (v) A test suite
verifies that every result file cited here exists and that the figure pipeline runs from stored
data alone. (vi) Results that changed a conclusion were re-derived independently before being
reported; two published claims of ours did not survive that process and were withdrawn.

\bibliography{refs}

\end{document}